# Refining the state-of-the-art in Machine Translation, optimizing NMT for the JA <-> EN language pair by leveraging personal domain expertise

Matthew Bieda
Supervised by: Julian Hough
Queen Mary University of London
MSc Computing & Information
Systems

*Abstract – Documenting the construction of an NMT (Neural Machine Translation) system for En/Ja based on the Transformer architecture leveraging the OpenNMT framework. A systematic exploration of corpora pre-processing, hyperparameter tuning and model architecture is carried out to obtain optimal performance. The system is evaluated using standard auto-evaluation metrics such as BLEU, and my subjective opinion as a Japanese linguist.*

I. INTRODUCTION

Effective communication between different groups is one of the oldest problems in human history. This has traditionally been addressed through the usage of a human being (interpreting) or the written word (translation). People have long dreamed of some perfect system that will enable flawless real-time global communication, but we are still far from this goal.

However, since the advent of computers, people have attempted to use them to solve this task. This concept of Machine Translation (MT) has evolved through 3 major paradigms thus far. Namely, they are Rule-Based Machine Translation (RBMT), Statistical Machine Translation (SMT) and finally Neural Machine Translation (NMT).

RBMT was the first organized attempt to use computers for the task of the translation around the 1950s, largely driven by the Cold War. It involves the creation of a bilingual dictionary and a set of grammar rules for each language to refer to during translation. In practice, RBMT was underwhelming, failing to produce fluent translations. Also, the initial cost in terms of funding and time to create these systems was very large. This led to the ALPAC report of 1966, which heavily criticized machine translation and led to a large reduction in US government funding (Poibeau, 2017).

SMT was pioneered by IBM in the 1990s. It involves the statistical analysis of parallel corpora to derive an approximated translation model (Brown, et al 1990). It has been rather successful and was the dominant approach in machine translation until the emergence of NMT in the last few years. Due to the inclusion of a monolingual language model which quantifies the likelihood of a translation, SMT produced more fluent translations than RBMT. Also, it did not require complicated linguistic rules which were expensive and time consuming to create. However, it did require lots of manual feature engineering to create the representative statistical models.

NMT has evolved commensurately with the Deep Learning revolution of the previous decade due to the increase in data availability and the decrease in GPU compute cost that has made the training of these deep sequence-to-sequence models viable. With NMT the complexity lies in the creation and optimization of the neural network architecture. Once established, all translation researchers must do is feed in appropriately pre-processed data and evaluate the output.

In the interests of clarification, I will cover some of the somewhat ambiguous terminology in the field here. Deep Learning is a subset of Machine Learning, which itself is a subset of Artificial Intelligence. The neural networks of Deep Learning take inspiration from the neurons of the human brain but are significantly less complex. The simplest neural network was the Perceptron model, a binary classification tool of the form $y=\sigma(w^Tx)$ where $\sigma$ is the Heaviside step function (a non-linear activation function).

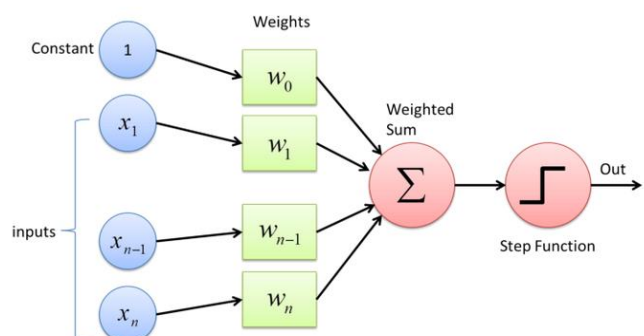

The Perceptron Model (Sharma, 2017)

This model is essentially Logistic Regression with a step function and was invented in 1958 by Frank Rosenblatt. This single-layer Perceptron could only be used to solve linearly separable classification tasks, famously being unable to learn even a simple XOR function. Researchers attempted to solve this problem with the creation of the Multilayer Perceptron (MLP), stacking multiple Perceptrons together.

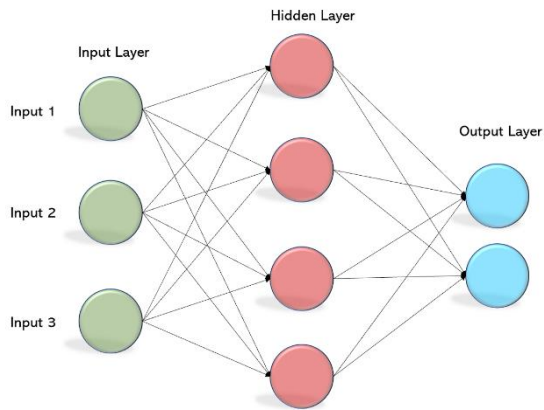

A more familiar network (Mohanty, 2019)

This architecture can achieve non-linear separation and solve the XOR classification problem, but the step function makes it difficult to train these models, as it does not preserve information between layers, and as optimally adjusting weights is provably NP-hard (Judd, 1988), solving complex problems is impractical.

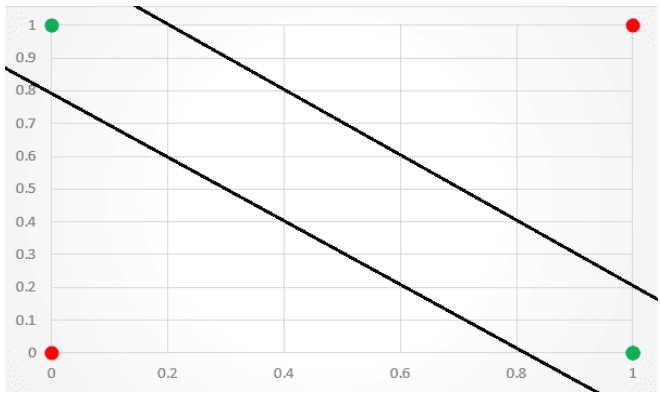

Classification solved with multiple decision boundaries

The breakthroughs that allowed the application of Perceptrons to more complex problems were the usage of a sigmoid activation function and the implementation of the backpropagation algorithm to automatically optimize weights. With these changes the basic building block of a neural network became known as a Neuron. The reason for the switch to the sigmoid activation is that backpropagation requires a continually differentiable function to work so the step function cannot be used. With this new paradigm neural networks could practically be used as universal function approximators. The only thing preventing mass adoption was the prohibitive computing power needed to train the models, which only became viable recently. These contemporary advances in compute allowed for the training of Deep Neural Networks (DNNs) with many hidden layers which have produced state-of-the-art results and have been the interest of much research discussed in the next section.

## II. RELATED WORK

Architecturally, the neural networks discussed so far are simple feed-forward neural networks (FFNNs). These can only take and output fixed-size vectors, but in the case of Machine Translation, neither the input nor the output can be thought of as fixed-size vectors. This is because every sentence passed into the system is going to be of variable length, as is the output generated in the other language. How can we solve this seemingly intractable problem?

We need an architectural change, namely the introduction of Recurrent Neural Networks (RNNs). RNNs allow us to solve sequence-to-sequence problems such as Machine Translation. It accomplishes this with the introduction of a looping mechanism that allows the network to contextualise the information from previous time steps.

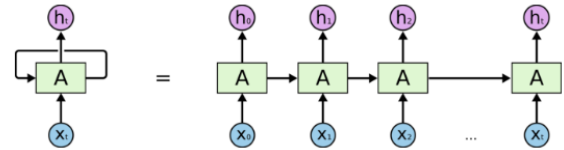

An unrolled recurrent neural network (Misra, 2019)

These RNN networks are practically implemented using the encoder-decoder model (Cho *et al*, 2014). Essentially this model uses one RNN to encode a variable length sequence into a fixed-size vector, which then acts as the input to the decoder RNN to transform back into a variable length sequence.

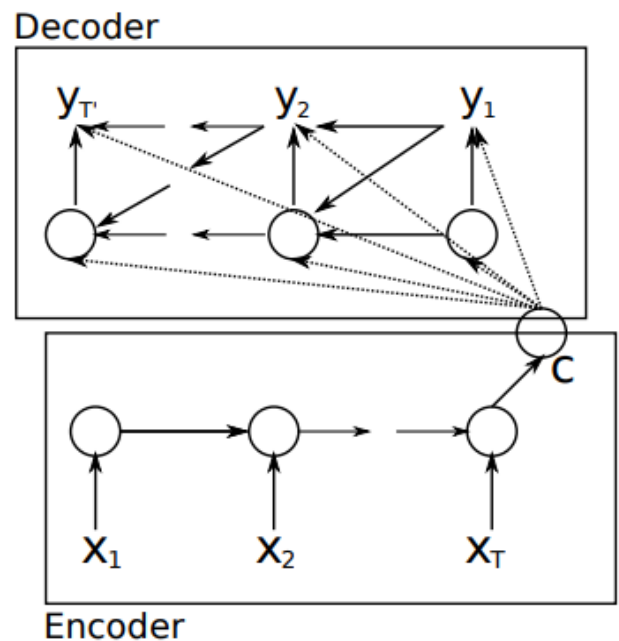

As conceptualized by (Cho, 2014)

However, this architecture is sub-optimal in several ways. It exhibits the vanishing gradient problem, a limitation of backpropagation in neural networks where gradients calculated deep in the network fail to propagate backwards and proportionately update earlier weights. Practically this significantly limits the sequence length that RNNs can deal with. Moreover, the compression of the entire input sentence into a fixed context vector is a lossy process in the case of long sentences, compromising accuracy. Also, this architecture has no capacity for parallelization, thus training speed is slow.

As the vanishing gradient problem is one of the most long-standing issues in neural networks research, many solutions exist. Generally changing the activation function to ReLU rather than the sigmoid or tanh activation is one way to ameliorate this problem. However, this is not as effective in the case of RNNs, instead we can use a modified RNN such as a Long Short-Term Memory (Hochreiter and Schmidhuber, 1997) or the Gated Recurrent Unit (Cho *et al*, 2014).

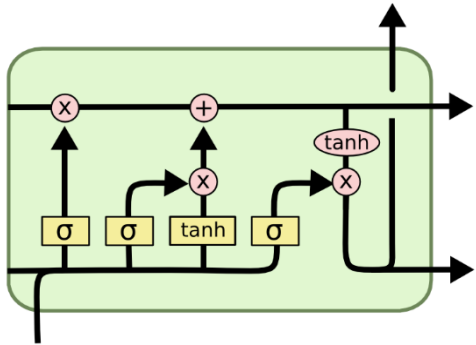

**An LSTM neuron (Olah, 2015)**

LSTM/GRU neurons contain a more complex internal structure that allows the network to filter out unnecessary information thereby augmenting the network's memory capacity and extending max sequence length.

The issue of the lossy context vector has been addressed by the introduction of the attention mechanism (Bahdanau, Cho and Bengio 2016), (Luong, Pham and Manning, 2015). Conceptually attention is a simple concept, but practically there are many different implementations, and it is a highly active research area. Essentially the attention mechanism allows the decoder side to leverage every hidden state of the input sentence, allowing it to attend to the most relevant parts of the input sequence when calculating its output. This extends the memory of the network significantly, and it also fixes the vanishing gradient problem by directly connecting the encoder and decoder. It is analogous to the skip connections utilized in Convolutional Neural Networks (CNNs).

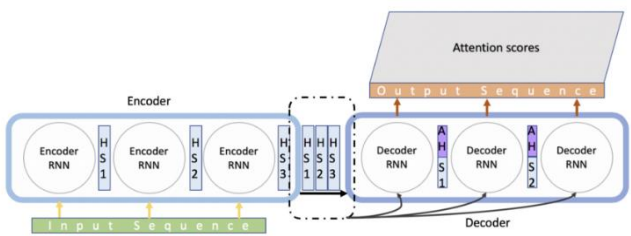

**The attention implementation in a seq2seq model (Dugar, 2019)**

The only remaining issue is that the inherently sequential nature of recurrent networks limits training speed. This was solved with the introduction of the Transformer architecture (Vaswani *et al.* 2017). The current state-of-the-art model in Machine Translation, the Transformer model built on top of the wave of research into attention mechanisms, resulting in a system that could perform set-to-set modelling while eschewing recurrence or convolution altogether. Hence the name of the paper in which it was introduced being "Attention is all you need". Because Transformers do not even use RNNs for encoding or decoding, they utilize a new "self-attention" mechanism to do so, inspired by the work of (Cheng, Dong and Lapata, 2016).

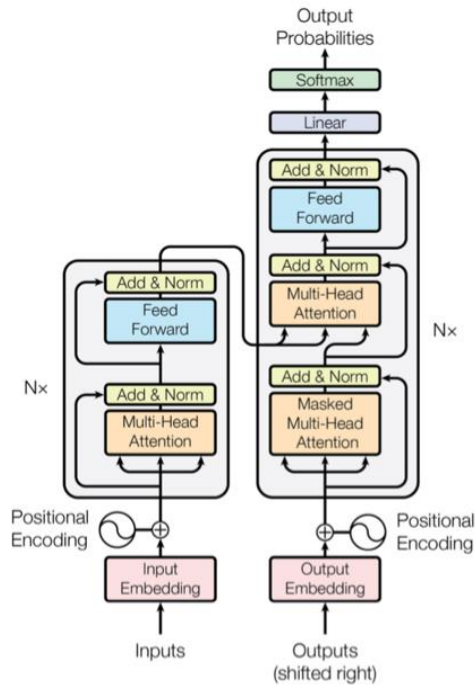

**Transformer architecture (Vaswani, 2017)**

A complete explanation of the Transformer is beyond the scope of this paper, but it is essentially an encoder-decoder model that leverages attention for the whole processing pipeline. Note that since the entire input sequence is now passed to the model simultaneously, we need a way to preserve some notion of sequence order. The method used here is to use layered sinusoidal waves to map positional encoding information. As of the Workshop for Machine Translation 2020, the current highest performing system for Japanese/English translation is from the NLP group at Tohoku University who utilized an ensemble Transformer model (Barrault *et al*, 2020).

Despite these architectural breakthroughs, preliminary NMT research struggled because of the OOV (out-of-vocabulary) problem. Since NMT systems dictate a fixed-vocabulary size of around 50,000 words, any rare words that cannot be included in this limited vocabulary will simply be marked by an unknown token: <unk>. This issue was first addressed by (Luong, 2015), who utilized a post-processing step that translated every OOV term using a dictionary, this increased BLEU scores by 2.8 compared to an equivalent system.

This was further improved upon by (Sennrich, Haddow and Birch, 2016), who used an implementation of the Byte-Pair

Encoding (BPE) algorithm to perform subword tokenization of rare words. BPE was originally conceived as a compression algorithm by Philip Gage in 1994. This was one of the less appreciated breakthroughs that catalyzed the development of famous language models such as BERT and GPT-3. This work was further extended (Wang, Cho and Gu, 2019) to apply BPE to Japanese and Chinese using byte-level subwords (BBPE). Also, another paper proposed Unigram, a modified subword regularization algorithm that leverages probabilistic sampling to choose subwords (Kudo, 2018). Recent research shows that Unigram language modelling provides superior results to BPE (Bostrom and Durrett, 2020). Also (Provilikov, Emelianenko and Voita, 2020), proposed a BPE-dropout implementation, that achieves a regularization effect like Unigram without having to use an alternative algorithm. This is the current state-of-the-art in subword tokenization, although note that this paper shows that Unigram is still superior for Japanese.

|  | BPE | Kudo (2018) | *BPE-dropout* |
|---|---|---|---|
| **IWSLT15** | | | |
| En-Vi | 31.78 | 32.43 | **33.27** |
| Vi-En | 30.83 | 32.36 | **32.99** |
| En-Zh | 20.48 | **23.01** | 22.84 |
| Zh-En | 19.72 | 21.10 | **21.45** |
| **IWSLT17** | | | |
| En-Fr | 39.37 | 39.45 | **40.02** |
| Fr-En | 38.18 | 38.88 | **39.39** |
| En-Ar | 13.89 | 14.43 | **15.05** |
| Ar-En | 31.90 | 32.80 | **33.72** |
| **WMT14** | | | |
| En-De | 27.41 | **27.82** | **28.01** |
| De-En | 32.69 | 33.65 | **34.19** |
| **ASPEC** | | | |
| En-Ja | 54.51 | **55.46** | 55.00 |
| Ja-En | 30.77 | **31.23** | **31.29** |

**Provilikov et al, 2020**

An aside about tokenization, the simplest way to tokenize text is to simply create one token for each word. But this creates far too many tokens, on the other hand simply creating a token for every character creates very few tokens but loses all the semantic meaning. Subword tokenization offers the best of both worlds but requires an algorithmic implementation.

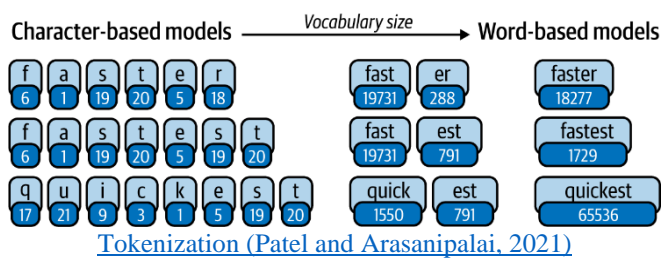

Tokenization (Patel and Arasanipalai, 2021)

Building good NMT systems is a multidisciplinary task, as it requires strong computer science skills and linguistic abilities. Luckily several frameworks have been created that make the process significantly simpler than if engineers had to begin from scratch. These include OpenNMT (Harvard, Systran), Fairseq (Facebook), Marian (Microsoft) and Sockeye (Amazon). These packages include efficient implementations of the Transformer and alternative architectures, pipelines for vocabulary generation & training, and guidelines for hyperparameter settings. Of these I will use OpenNMT (specifically the TensorFlow version) as it offers many customization options, ease of use for non-machine-learning specialists and an active community forum where researchers reply to any queries.

For implementing Subword tokenization, the SentencePiece package (Google) offers support for BPE, BBPE, BPE-dropout and Unigram models, as well as naïve character and word encoding. The Subword-NMT package (Sennrich – University of Zurich) is less feature rich and easy to use. Note that WordPiece is a Google internal implementation of SentencePiece to train BERT that utilizes a modified BPE algorithm.

III. METHODOLOGY

**Naïve Experimentation**

Based on the above research, my implementation of a Japanese/English Neural Machine Translation system will be based on the Transformer architecture to achieve the best accuracy. Many open-source implementations exist online in projects such as Marian, Fairseq and OpenNMT. My research will use OpenNMT in TensorFlow as it is a popular framework and has the most active research community. The operating environment will be a Google Colab Pro instance, using an Nvidia P100/V100 GPU and an Intel Xeon CPU. I chose this service because it offers an extremely large amount of computing power for £8.10 monthly. I also purchased up to 2 TB of storage space on Google Drive to be able to synchronize all my model checkpoints for £7.99 a month.

Like any AI system, the quality of an NMT model is largely dependent on the quality and amount of the training data. The most comprehensive list of parallel corpora for the Japanese / English language pair is maintained by Professor Graham Neubig at http://www.phontron.com/japanese-translation-data.php. Of these I have selected JParaCrawl, Japanese-English Subtitle Corpus (JESC), ASPEC, Kyoto Corpus, JaEn Legal Corpus and TED talks as they are high-quality and can easily be pre-processed into the requisite format. In total this is 15,847,858 sentences parallel aligned. Note that the ASPEC corpus was received as three files containing 1 million sentences each, but documentation mentioned that the last of these was of significantly lower quality, so I only included the first two million. Some of these datasets required significant pre-processing to separate the sentences contained into a separate file for each language. This was accomplished using the regex module in Python as well as the AWK programming language and SED for stripping leading whitespace.

The corpora used in this study are all high-quality, but some are imperfect. Research shows that cleaning of parallel corpora is an effective way to increase accuracy (Ramírez‑Sánchez *et al*, 2020) states: "Cleaned corpora, in the most aggressive cleaning scenario (Bicleaner scores above 0.7), lead consistently to equal or slightly better results for BLEU scores in machine translation, not degrading the results in any case and speeding up machine translation training times." Some of the corpora I have used were pre-scored, so it is simple to establish some threshold for cleaning. However, to establish a baseline, the first experiment will be performed without cleaning or any form of tokenization.

Some corpora were only available as a single file, i.e., they were not pre-split into train, validation, and test sets. The purpose of the validation set is to ensure that the model does not overfit on the training data, and the test set is used to validate the quality of the model on unseen data. Therefore, it was required to split these files into the various data sets. This was accomplished with a script available at: https://github.com/ymoslem/MT-Preparation. Note that I initially tried to run this on my local machine (16GB RAM) but ran into an out of memory error. Subsequently I ran it inside a "High RAM" Colab environment (25.46 GB) which was sufficient for my largest files.

My experimental design was to increase complexity over time and observe the effects on the output of the model. So, for the baseline test I simply took the largest corpora (JParaCrawl – 10,120,013) and split the English and Japanese segments into separate files. Then I ran the script mentioned above to create train, validation, and test sets. Next, I tuned the OpenNMT configuration file to use the Transformer architecture with default hyperparameter options. After installing OpenNMT in the Colab environment, I computed the vocabulary over the whole corpora as recommended in the documentation. Then I simply trained the model using default arguments.

The problem I ran into is because I had not pre-processed my corpora, the vocabulary construction stage generated an extremely large number of tokens for each language. Specifically, there were 4,237,855 English tokens and 12,763,071 Japanese tokens. By default, OpenNMT cuts this down to 50,000 on each side by frequency because of the prohibitively large RAM and VRAM requirements of significantly higher thresholds.

Upon translating the test set using the model with lowest validation perplexity of around 18, the results demonstrated well the Out of Vocabulary (OOV) issue with naïve NMT. Most sentences were simply outputted as the <unk> token, for prediction and reference. Of the sentences that were intelligibly parsed, some of the standouts were the following:

Test: "If you have questions about specific retention periods, please contact us via the contact details included below".

Model output: 特定(specific)の保持期間 (retention period) について(about)ご質問(inquiry)があり場合 (if you have)、下記 (the following) に記載されている(listed)連絡先 (contact address) よりお問い合わせください。(Please enquire to).

Reference: 特定(specific)の Cookie の有効期間 (validity period) について(about)ご質問(inquiry)など(etc)がありましたら (if you have)、弊社まで (to our company) お問い合わせください (please enquire) (連絡先は以下に記載)。 (Contact info listed).

This is a great translation; it is completely comprehensible and unambiguous. The only issues are minor grammatical errors, specifically あり場合 should be the unconjugated verb form ある場合, and the より particle should be replaced with に. It is a more literal interpretation than the reference as the model does not have access to the surrounding context.

Test: "If the UPDATE feature was disabled, the Asus K53TA Notebook LiteOn WLAN Driver 9.0.0.223 for Windows 7 x64 driver could not be installed.
Model output: アップデート"機能がオフになっていると、Asus <unk> Notebook ATKOSD2 Utility <unk>
Reference: アップデート"機能がオフになっていると、Asus K53TA Notebook ATK ACPI Driver 1.0.0014 for Windows 7 x64 ドライバは正しくインストールされません。

This example shows the effects of having such a limited vocabulary. The model was unable to deal with the specifics of the computer system as it was out-of-vocabulary. One obvious way to fix this problem is to simply increase the size of the vocabularies that the model is trained on. However even at 250k I was running into memory errors, although I managed to train a model successfully at 200k. This achieved a superior validation perplexity of 11.2. But when attempting to use this model to translate the test set, I encountered another CUDA memory error and could not process the whole set. While I could see an improvement in these results, it was still far from being useable.

Another option to test is the ability to share vocabulary between the languages. Normally this is a function reserved for closely related languages, but since the sentences in my corpora kept some terms untranslated it may be effective here.

## **Comprehensive Pre-Processing**

Clearly, we need another layer of complexity to address this problem. The answer lies in pre-tokenizing the corpora so that a meaningful vocabulary can be built. We do not want to use a naïve word tokenization algorithm delimited on space as this will generate a large vocabulary. Instead, we will utilize the state-of-the-art subword tokenization algorithms, specifically Unigram. The open-source SentencePiece project provides a language agnostic end-to-end solution for implementing subword tokenization algorithms in NMT systems. There is a complete pre-processing script collection available at https://github.com/ymoslem/MT-Preparation.

For this step I also concatenated together my 6 datasets to have the largest possible parallel corpora that covers several domains. I then filtered and then tokenized my dataset using the Moses tokenizer for both source and target. Based on these Google SentencePiece experiments: https://github.com/google/sentencepiece/blob/master/doc/experiments.md, the best English to Japanese results were obtained by pre-tokenizing with Moses for English and MeCab for Japanese, then passing through a SentencePiece model. When training a SentencePiece model vocab size is an important hyperparameter if Japanese is the target.

```
Dataframe shape (rows, columns): (10120013, 2)
--- Rows with Empty Cells Deleted       --> Rows: 10120013
--- Tokenizing the Source Complete      --> Rows: 10120013
--- Tokenizing the Target Complete      --> Rows: 10120013
--- Duplicates Deleted                  --> Rows: 8800926
--- Source-Copied Rows Deleted          --> Rows: 8800780
--- Too-Long Source/Target Deleted      --> Rows: 3350814
--- Rows with Empty Cells Deleted       --> Rows: 3350814
```

**Results of filtering the corpora.**

Note that many sentences were removed here when checking for a discrepancy between source and target length. This may have been tuned too aggressively for the English/Japanese pair, as the equivalent Japanese sentence tends to be significantly shorter in terms of characters. This could be adjusted to be less aggressive and leave a larger corpora size. However ultimately, I decided that this filtering was unnecessary and was over-normalizing my data, so I simply used pre-tokenization alone.

Following this pre-processing, I then split the corpora into train, validation, and test files before training the SentencePiece model and generating the vocabulary files as to not bias the data. I used 5000 lines each for the validation and test files.

Next, I trained a SentencePiece Unigram (32k vocab size) model on this tokenized training data. Note that the Colab instance with 25GB of RAM failed here, even with the optional "-train_extremely_large_corpus" flag enabled. So, I had to complete this section using an Amazon Web Services EC2 instance.

The SentencePiece configuration also supports the ability to decompose unknown pieces into UTF-8 byte-level fragments to further alleviate the OOV problem. This is essentially an implementation of the Byte-Level BPE (BBPE) which was shown to be advantageous for Asian languages, therefore I will also try this in another experiment. Now finally we can process the subword regularization of the corpora using the SentencePiece models created previously. This generates the subword encoded data files we can pass to the NMT system. Lastly, we must convert the vocabulary files constructed by SentencePiece during model creation into valid OpenNMT-tf files.

Then we can simply use these vocabulary files and the subword encoded training data to train the transformer model. Due to time constraints all models were trained for 40,000 steps. Then finally we can pass the test data into our trained model and generate the attempted translations. This workflow produces much higher quality results. Most input sentences from the test set have received an accurate predicted translation. Now that we have a good output, we can continue to the automatic evaluation step. To perform this, firstly we must decode the test set and the prediction output to remove the subword regularization. Then if you performed an initial pre-tokenization step, we must first detokenize to return to the original text before passing it to be evaluated by SacreBLEU. This is necessary because it employs its own metric-internal tokenization, allowing for more meaningful comparisons between papers, as explained by (Post, 2018).

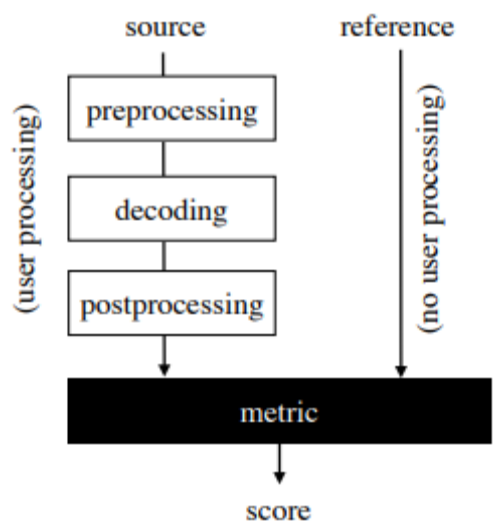

Desired workflow for reproducible BLEU scores (Post, 2018)

SacreBLEU is a software package that implements the BLEU evaluation metric in a reliable and reproducible manner. Proposed by IBM, BLEU evaluates the correspondence between a reference translation and the machine's attempt by matching n-grams (Papineni *et al*, 2002). Although it is not perfect, it tracks well with human judgment and is very popular in this research space. It is scored between 0 and 100, but realistically anything over 40 is likely to be a high-quality translation.

Note that the default internal tokenizer of SacreBLEU is 13a which is an implementation of the mteval-v13a Perl script from the Moses project. For Japanese we will want to make sure to use the alternative MeCab tokenizer, otherwise reported scores will be far lower than they should be. On my internal test set I received a score of 32.7. However ideally, we should be using a common reference test set, so the performance of the NMT system can be compared properly even if other researchers are using different training and validation data. Since my system is not domain specific, I opted for the Workshop for Machine Translation 2020 official evaluation data, which is sourced from news media. My system achieved a score of 19.7 on this reference test set.

For context Google Translate achieved 24.2, so my system is relatively close to the state-of-the-art software.
Note that all my experiments so far were translating into Japanese. If you are translating into English, you must make sure to run the output through a true-casing script as the model can only output lowercased data. This is the most efficient way of maintaining case information without losing accuracy. Finally, I performed more experiments to find my optimal model. I experimented with BBPE and pre-tokenization.

## 2022 Update

I ran several more experiments. I Identified a tendency of the system to output less "Sino-Japanese" words in favor of more natural spoken Japanese expressions, giving the system a more human quality. However, this is something that BLEU would penalize as the reference test used the former, although the semantic meaning is the same. This demonstrates the limitations of automatic evaluation systems. I also experimented with pseudo document-level analysis, i.e., simply inserting a space between every new corpus in the training data as well as separate pre-tokenization pathways and model averaging. I employed all of these for my latter experiments. I also simply scaled up the system by using a larger vocabulary and Transformer. I have deployed my best model on the web at jp-translate.cloud

```
 1  line 59
 2  府のEU離脱政策が北アイルランドの和平交渉を損なうとした異議申し立ては、高等法院により先に棄却されていた。
 3  政府のブレクジット政策が北アイルランド和平プロセスに損害を与えると主張したものは、以前に高等裁判所によって退けられました。[mine]
 4
 5
 6  line 103
 7  ただし、外国人女性は「控えめな服装」が求められる、と詳細説明は抜きで同氏は補足した。
 8  しかし、外国人女性は「控えめな服」を着なければならないと彼は付け加えた。 [mine]
 9
10  line 114
11  サウジアラビアは、ゼロから観光産業を構築しようと巨額の資金を散財している。
12  サウジアラビアは、ゼロから観光産業を構築するための試みで数十億を費やしています。 [mine]
```

Demonstrating the similar semantics but less formal register of my model vs Google.

IV. RESULTS

| | Validation Perplexity | BLEU Score (internal data) | BLEU Score (WMT data) | Subjective Analysis |
|---|---|---|---|---|
| Raw Data (JParaCrawl only) – No pretok | 18 | N/A | N/A | Unusable due to <unk> tokens |
| All Data – Moses Pretok only | 3.5 (artificially low) | N/A | N/A | Unusable due to <unk> tokens |
| All Data – Moses Pretok + SentencePiece (Unigram-32k) | 7.61 | 34.4 | 18.4 | Useable for non-professional contexts |
| All Data – Moses Pretok + SentencePiece (UnigramBBPE-32k) | 7.63 | 34.7 | 17.9 | Useable for non-professional contexts |
| All Data – Moses Pretok + SentencePiece (UnigramBBPE-32k English 8k Japanese) | 5.26 | 33.3 | 18.0 | Marginal difference with smaller Japanese vocabulary |
| All Data – Moses Pretok + SentencePiece (Unigram-32k) Document level corpora | 7.39 | 34.1 | 18.0 | Marginal difference with pseudo Document-level analysis |
| All Data – Moses & Mecab Pretok + SentencePiece (Unigram-32k) | 4.66 | 34.7 | 18.9 | **Best model** |
| Model averaging for above | N/A | 34.8 | 18.9 | Slightly better with averaging |
| Best model – Trained to convergence, increased vocab size (100k each), used BBPE and "split_digits", and using Big Relative Transformer and averaged last 5 | 3.71 | 38.6 | 19.7 | Significant improvements seen with these methods, although much more computationally expensive. Useable for professional contexts |
| Above but into English | 4.44 | 37.1 | 25.0 | Great performance also in reverse direction |

We can see here that subword tokenization is not optional in the construction of a functional NMT system. Even with a pre-tokenization step the OOV issue is too severe. Note that we can completely fix this OOV problem with BBPE, as all rare words will be broken down into their constituent bytes. In a standard subwording implementation like Unigram, if a rare word cannot be constructed from subwords it must be declared OOV. BBPE also benefits transfer learning and multilingual modelling.

Using a larger transformer and vocabulary size I was able to increase performance significantly, but these models took multiple weeks to train on a fast GPU.

## V. CONCLUSION

In this paper I have documented my attempt to create a high-quality NMT system as a newcomer to this field. Therefore, I have also addressed the history of machine translation research, as well as the architectural details of modern NMT systems. Through a discussion of available NMT frameworks and tools, and a detailed record of my workflow, I hope that this paper can serve as a reference to other researchers learning to create NMT systems.

Based on the above exploration, I have constructed some guidelines for the development of NMT systems in general. Note that although some of these may seem obvious to computer scientists, they may not be so for those with a linguistics or languages background.

- Amass a large amount of parallel data, ideally at least several million lines worth. If you are working with a low-resource language pair, consider data augmentation techniques such as back translation. These are surprisingly effective.
- Ensure your training data is of high quality, make sure to clean your data well and ensure any nonsense characters or misaligned translations are removed. Quality of data is the most important factor after quantity. This process can be largely automated through scripts if manual checks would be too time consuming.
- Subword regularization algorithms are mandatory for functional systems. However, note that training SentencePiece models on large datasets requires prohibitively large amounts of RAM. Either complete training through a high-RAM EC2 instance or use a shuffled sample of your sentences for training. BBPE should provide a particular improvement for character-based languages.
- Utilize the Transformer architecture, it allows for far more accurate results compared to previous models. There is no need to write your own implementation from scratch, many tried and tested versions are available in frameworks such as OpenNMT, Fairseq and Marian. Once you have tested the base transformer, try some of the modified architectures supported in the framework.
- Cloud based services such as Google Colab provide an extremely cheap way of training large deep learning models quickly with enterprise hardware. For extremely RAM intensive training use an AWS EC2 instance.
- Automatic evaluation metrics such as BLEU are a good way to benchmark your results, but subjective analysis by a language expert remains the gold standard. Language expertise will also allow you to fine tune your system quicker and more optimally, as you will understand how and why various hyperparameter adjustments will impact your results, and the idiosyncrasies of the languages used. If using BLEU, ideally use a reference test set rather than an internal one.
- Optimizations you can employ include changing sub-wording algorithms, employing BBPE, vocabulary size adjustments, sharing embeddings, ensembling models and using a bespoke tokenizer if one exists for your use case.

## VI. FUTURE WORK

I hope this work can be a valuable contribution to the translation and FOSS communities. Japanese ←→ English is one of the most requested language pairs, but human translators who are experts in both languages are few and far between. Improved performance could probably be obtained by using a cleaned version of JParaCrawl, as it was published with a Bicleaner cutoff of 0.5, studies show 0.7 is a better cutoff as NMT is very sensitive to noise. Also shared embeddings are worth testing, even though they are generally used for similar languages.

Probably the most significant breakthrough that NMT currently needs is the introduction of document/paragraph level context. Currently even the most advanced systems are limited to sentence-level context. However, research by Castilho (2021) showed that in English to Portuguese, over 33% of sentences could not be adequately translated with only sentence level context. Assuming this generalizes to other language pairs, NMT systems will be unable to match professional translation until this limitation is solved.

Thankfully NMT research is a popular field, and many ingenious solutions have been proposed to solve existing limitations. For example, a novel approach to share information between the embedding layers (Liu *et al*, 2019), and architectural modifications that normalize decoding accuracy across the whole output sentence (Zhou, Zhang and Zong, 2019). However, these solutions tend to be rather complex, and if they are not validated by other researchers and subsequently integrated into production frameworks, they are unlikely to be widely adopted. Researchers will likely continue to work on optimizing sentence-level context first, then tackle the hard problem of achieving wider contextualization.